\documentclass[runningheads]{llncs}

\usepackage{eccv}

\usepackage{eccvabbrv}

\usepackage{graphicx}
\usepackage{booktabs}
\usepackage{lipsum}
\usepackage{colortbl}

\usepackage[accsupp]{axessibility}  % Improves PDF readability for those with disabilities.

\usepackage{hyperref}

\usepackage{orcidlink}
\usepackage{comment}
\usepackage{multirow}
\usepackage{wrapfig} 
\usepackage{makecell}
\usepackage{caption}
\usepackage{blindtext}
\usepackage{float}

\usepackage{etoolbox}
\AtBeginEnvironment{tabular}{\scriptsize}

\begin{document}
\title{DNI: Dilutional Noise Initialization for Diffusion Video Editing}
\titlerunning{DNI: Dilutional Noise Initialization for Diffusion Video Editing}
\author{Sunjae Yoon\orcidlink{0000-0001-7458-5273} \and
Gwanhyeong Koo\orcidlink{0009-0005-6455-3223} \and
Ji Woo Hong\orcidlink{0000-0002-3758-0307} \and 
Chang D. Yoo\orcidlink{0000-0002-0756-7179}}
\authorrunning{S. Yoon et al.}
\institute{Korea Advanced Institute of Science and Technology, Daejeon, Republic of Korea
\email{\{sunjae.yoon,cd\_yoo\}@kaist.ac.kr}\\
}
\maketitle
\begin{figure}[h]
\begin{center}
\centerline{\includegraphics[width=0.9\textwidth]{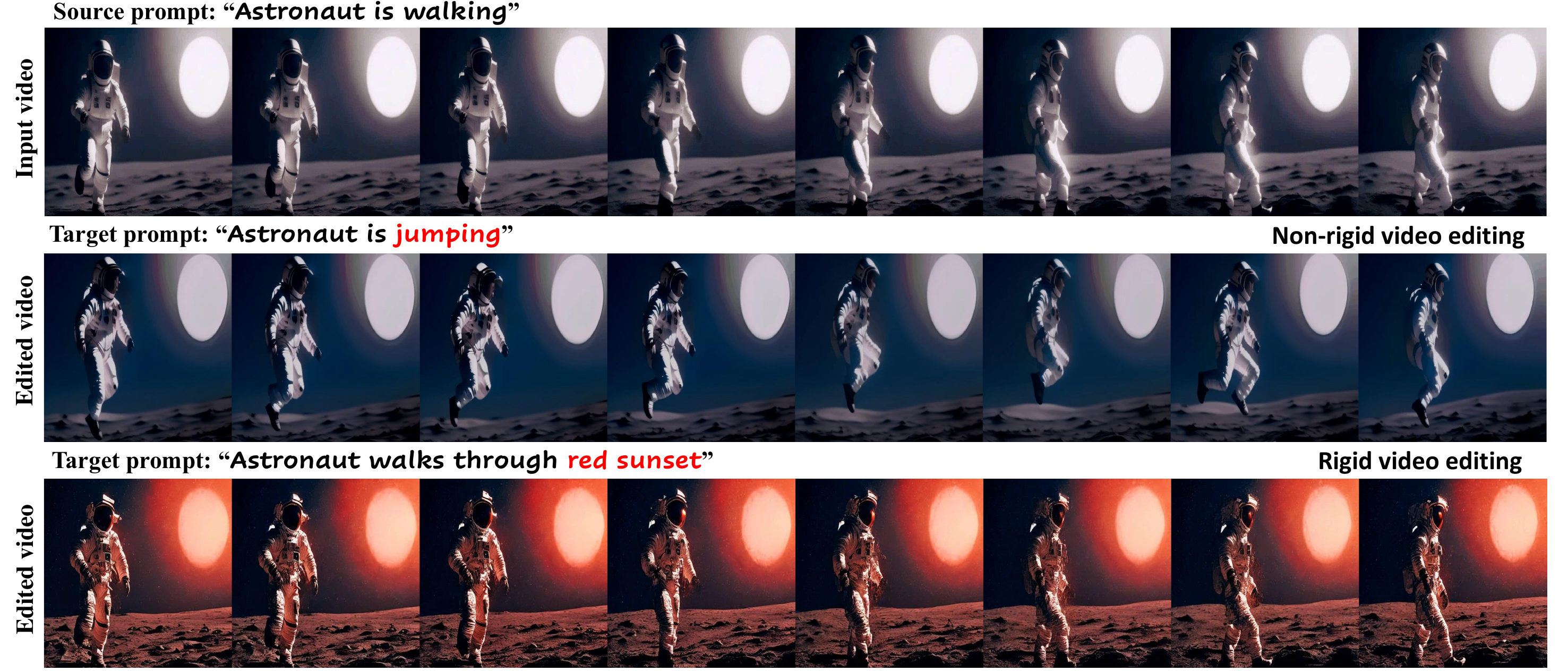}}
\caption{Edited videos of Dilutional Noise Initialization (DNI) framework. DNI performs text-based rigid and non-rigid edits, enabling effective alteration under high fidelity.}
\label{fig:teaser}
\end{center}
\end{figure}
\begin{abstract}
  Text-based diffusion video editing systems have been successful in performing edits with high fidelity and textual alignment. However, this success is limited to rigid-type editing such as style transfer and object overlay, while preserving the original structure of the input video. This limitation stems from an initial latent noise employed in diffusion video editing systems. The diffusion video editing systems prepare initial latent noise to edit by gradually infusing Gaussian noise onto the input video. However, we observed that the visual structure of the input video still persists within this initial latent noise, thereby restricting non-rigid editing such as motion change necessitating structural modifications. To this end, this paper proposes Dilutional Noise Initialization (DNI) framework which enables editing systems to perform precise and dynamic modification including non-rigid editing. DNI introduces a concept of `noise dilution' which adds further noise to the latent noise in the region to be edited to soften the structural rigidity imposed by input video, resulting in more effective edits closer to the target prompt. Extensive experiments demonstrate the effectiveness of the DNI framework. 
\end{abstract}

\begin{figure}[t!]
\centering
    \includegraphics[width=1.0\textwidth]{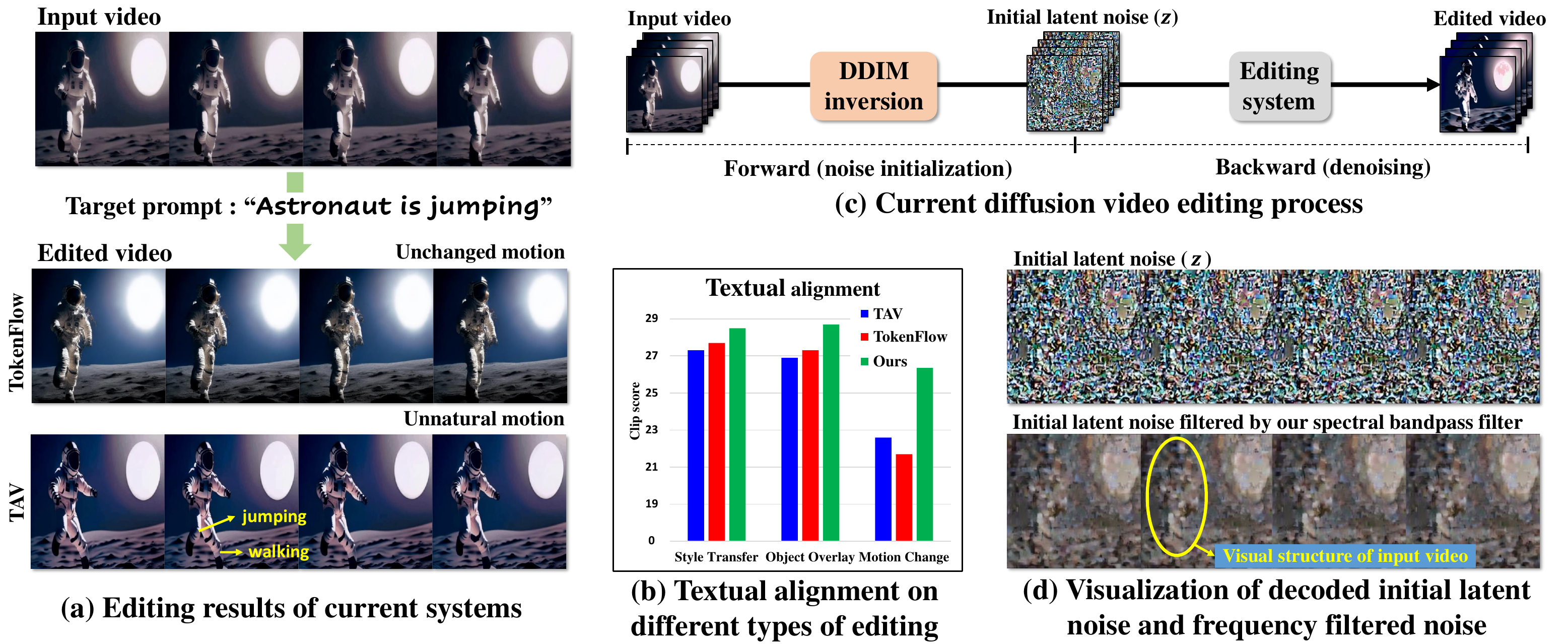}
   \caption{(a) Editing results about motion change of current systems \cite{geyer2023tokenflow,wu2022tune}. (b) Categorical analysis of textual alignment with video across different types of editing on DAVIS \cite{pont20172017}. (c) Overview of current diffusion video editing process. (d) Visualization of initial latent noise and the latent noise filtered by our designed adaptive spectral filter, where the input video's visual structure clearly remains in the initial latent noise.}
\label{fig:introduction_1}
\end{figure}
\section{Introduction}
\label{sec:intro}
Denoising diffusion models \cite{dhariwal2021diffusion,song2020score,song2020denoising,ho2020denoising} have spurred substantial innovations in the generative capabilities of artificial intelligence. 
By gradually denoising input Gaussian noise, diffusion models generate various outputs including image \cite{rombach2022high,nichol2021glide}, audio \cite{kong2020diffwave,yang2023diffsound}, and video \cite{ho2022imagen,chen2024videocrafter2}, which can be further edited to meet users' specific needs.
We focus here on diffusion video editing which holds immense promise for revolutionizing the entertainment industry.
Video editing systems aim to modify specific attributes in an input video corresponding to users’ requirements from a target textual prompt.
Recent video editing systems \cite{wu2023tune,liu2023video,qi2023fatezero,geyer2023tokenflow} have succeeded in performing edits with high fidelity to input video and precise textual alignment.
However, this success is still restricted to rigid modifications such as style transfer and object overlay by preserving input video structural layouts.
Specifically, in \cref{fig:introduction_1} (a), for a target prompt requiring non-rigid modifications (\eg ``Astronaut is jumping''), current systems fail to conform and return the original input video under over-fidelity.
Otherwise, they often exhibit unnatural motion by blending the original content (\ie walking) with the targeted content (\ie jumping) in the video.
In \cref{fig:introduction_1} (b), our categorical analysis of textual alignment with video across different editing types (\ie motion change, style transfer, object overlay) demonstrates that current systems are struggling with the motion change.
Therefore the resulting videos about complex non-rigid editing still remain unsatisfactory.
Our investigation revealed that one of the reasons for this unsatisfactory non-rigid editing stems from the initial latent noise fed into the diffusion video editing systems.
In a typical process of diffusion video editing, as shown in \cref{fig:introduction_1} (c), the diffusion model infuses a Gaussian noise onto input video to build initial latent noise $z$ using inverse denoising (\eg DDIM inversion) as a forward process.
In the backward process, the model performs denoising into this $z$ to generate edited videos conforming to the target prompt.
Despite the noise $z$ assuming a Gaussian noise distribution, in \cref{fig:introduction_1} (d), we observe that this noise still contains the visual structures of the input video.
To ascertain their presence, we devise a frequency pass filter referred to as adaptive spectral filter, which captures the clearer structure of the input video within the latent noise.
Consequently, the current video editing systems perform edits on top of the input video's visual structure within the initial latent noise, facilitating rigid editing yet exposing a susceptibility to non-rigid editing that necessitates altering the structure.
\begin{figure}[t!]
    \centering   
    \includegraphics[width=1.0\textwidth]{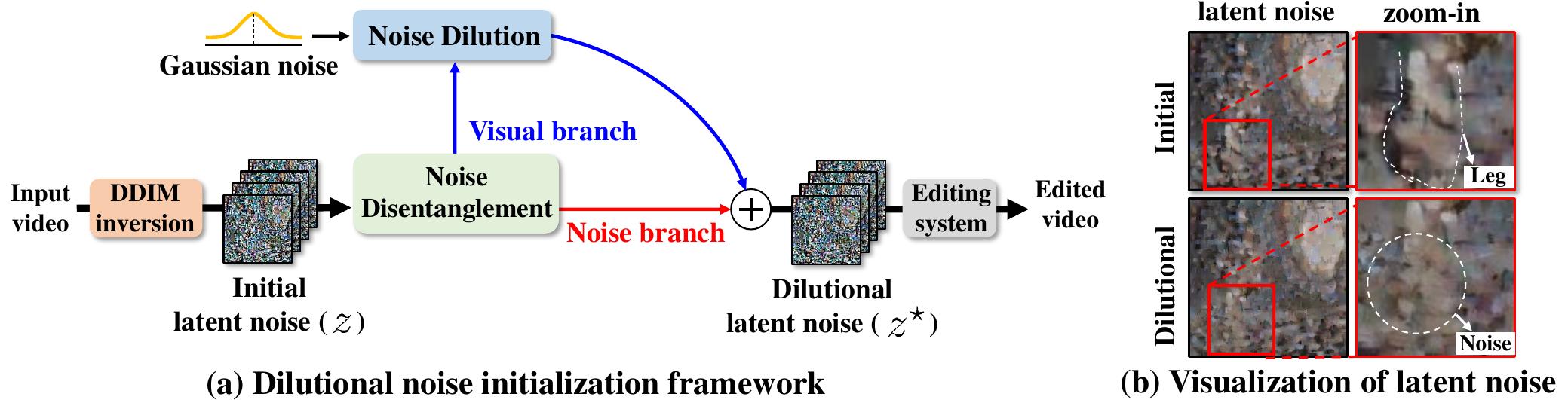}
    \caption{(a) Illustration of Dilutional Noise Initialization framework. The noise disentanglement separates the initial latent noise into a visual branch and a noise branch. The visual branch contains a visual noise of input video components and the noise branch contains a Gaussian noise. The noise dilution adds further noise into an editing region of the visual noise, enabling dynamic modifications without being restricted by the input video layout. (b) Visualizations of initial and dilutional latent noises.}
    \label{fig:introduction_2}
\end{figure}
To this end, we propose Dilutional Noise Initialization (DNI) that enables video editing systems to perform precise and dynamic modifications encompassing the versatility of non-rigid editing.
As shown in \cref{fig:introduction_2} (a), the DNI framework introduces a novel concept of \textit{noise dilution}, which adds further noise into latent noise to ensure the edited video aligns more closely with the input text prompt.
Formally, the DNI takes an initial latent noise $z$ as input and produces a dilutional latent noise $z^{\star}$ which mitigates structural rigidity imposed by the input video in the area to be edited.
To build the $z^{\star}$, DNI framework performs two main processes: (1) disentangling initial noise into the visual branch and noise branch and (2) diluting the noise in the visual branch with additional Gaussian noise.
For the noise disentangling, we design a frequency pass filter referred to as \textit{adaptive spectral filter} which effectively isolates the input video components into the visual branch by considering the frequency spectrum of the input video.
Subsequently, dilution is carried out within the visual noise by blending Gaussian noise into the targeted editing area guided by the target prompt.
Finally, dilutional latent noise $z^{\star}$ is synthesized by recombining the noises in the two branches.
As shown in \cref{fig:introduction_2} (b), the dilutional latent noise preserves the input video's visual structure while simultaneously reducing the rigidity of this structure in the specific area targeted for editing (\eg the man's legs).
DNI is applied to any diffusion video editing system in a model-agnostic manner, demonstrating effective editability on video editing benchmarks (DAVIS \cite{pont20172017}, TVGVE \cite{wu2023cvpr}).
%
%-------------------------------------------------------------------------
\section{Related Works}
\label{sec:formatting}
\subsection{Diffusion-based generative models}
Denoising diffusion models \cite{ho2020denoising,song2020denoising} have surpassed erstwhile qualities of generative adversarial networks \cite{goodfellow2020generative}.
% Deep diffusion model의 등장 및 GAN을 앞도함.
%
Diffusion-based text-to-image (T2I) models \cite{saharia2022photorealistic,ramesh2022hierarchical} have significantly advanced image generation, producing high-fidelity images from text. 
These models are now extending into text-to-video (T2V) task.
Early works \cite{wu2022nuwa,ho2022video,hong2022cogvideo} in T2V task adapted pre-trained T2I models by incorporating a temporal dimension, where many temporal attentions \cite{ho2022imagen,singer2022make} are also designed to enhance frame consistency.
Recently, diffusion models have excelled in various generative works including super-resolution and inpainting \cite{saharia2022image,lugmayr2022repaint}. 
Among these, diffusion video editing presents a new challenge for controlled synthesis across frames obtrusively, discussed in detail below.

\subsection{Diffusion video editing }
The recent success of text-based image editing \cite{hertz2022prompt,brooks2022instructpix2pix,koo2024wavelet} bridges to video editing \cite{wu2022tune,geyer2023tokenflow,yoon2024frag}.
For video, the challenge lied in seamlessly integrating edited frames under high fidelity to input video.
Thus, several technical solutions to enhance temporal consistency are introduced including temporal attention \cite{wu2022tune,geyer2023tokenflow} and knowledge injection \cite{liu2023video,qi2023fatezero} from input video priors.
Previous models sought to preserve the input video's information to improve editing quality \cite{song2020denoising}, yet paradoxically, they encountered a trade-off, sacrificing versatility in editing.
To this end, we present a DNI framework that can enable video editing systems to perform various edits including non-rigid edits, maintaining their video quality.

\section{Preliminaries}
\label{sec:prelim}
\subsection{Denoising diffusion probabilistic models}
\label{gen_inst}
Denoising diffusion probabilistic models (DDPMs) \cite{ho2020denoising} are structured as parameterized Markov chains, methodically restoring noisy data sequences $\{x_{1}$,$\cdots$, $x_{T}\}$ from an initial $x_{0}$.
First, Gaussian noise is progressively added to $x_{T}$ through the Markov transition $q(x_{t}|x_{t-1}) = \mathcal{N}(x_{t};\sqrt{\alpha_{t}}x_{t-1},(1-\alpha_{t})I)$, following a predefined schedule $\alpha_{t}$ over steps $t\in\{1,\cdots,T\}$.
This procedure is defined as the \textit{forward process} in diffusion modeling.
The \textit{reverse process} is then applied to generate data using diffusion model estimating $q(x_{t-1}|x_{t})$ through trainable Gaussian transitions $p_{\theta}(x_{t-1}|x_{t}) = \mathcal{N}(x_{t-1};\mu_{\theta}(x_{t},t),\sigma_{\theta}(x_{t},t))$, starting from the normal distribution $p(x_{T}) = \mathcal{N}(x_{T};0,I)$.
The model is trained to maximize log-likelihood $\textrm{log}(p_{\theta}(x_{0}))$ over $\theta$, where variational inference maximizes the lower bound of $\textrm{log}(p_{\theta}(x_{
0}))$, yielding a closed-form KL divergence between distributions $p_{\theta}$ and $q$.
This process is summarized as training a denoising network $\epsilon_{\theta}(x_{t}, t)$ to predict noise $\epsilon \sim \mathcal{N}(0,I)$ as $\mathbb{E}_{x,\epsilon \sim \mathcal{N}(0,1),t \sim \mathcal{U}\{1,T\}}[||\epsilon - \epsilon_{\theta}(x_{t},t)||_{2}^{2}]$,
where $\mathcal{U}\{1,T\}$ is discrete uniform distribution from 1 to $T$ for robust training on each step $t$.
\subsection{Denoising diffusion implicit model and Inversion}
Denoising diffusion implicit model (DDIM) \cite{song2020denoising} accelerates diffusion reverse process, sampling with fewer steps as 
$x_{t-1} = \sqrt{\frac{\alpha_{t-1}}{\alpha_{t}}}x_{t} + \left( \sqrt{\frac{1}{\alpha_{t-1}}-1} - \sqrt{\frac{1}{\alpha_{t}}-1} \right) \epsilon$.
We can also build an inverse process of this acting as the forward process as $x_{t+1} = \sqrt{\frac{\alpha_{t+1}}{\alpha_{t}}}x_{t} + \left( \sqrt{\frac{1}{\alpha_{t+1}} - 1} - \sqrt{\frac{1}{\alpha_{t}} - 1} \right) \epsilon$, referred to as DDIM inversion process. 
In diffusion editing, DDIM inversion enhances fidelity to the input video.
\subsection{Text-conditioned diffusion model}
\label{sec:3.2}
The text-conditioned diffusion model generates the output data $x_{0}$ conditioned on a text prompt.
The training objective incorporates textual condition under latent space as $\mathbb{E}_{z,\epsilon,t}[||\epsilon - \epsilon_{\theta}(z_{t},t,\mathbf{c})||_{2}^{2}]$, where $z_{t}$ is a latent noise encoding of $x_{t}$ using VQ-VAE \cite{van2017neural} and $\mathbf{c}$ is target prompt CLIP \cite{radford2021learning} embedding.
Video editing takes $z_{t}$ as input video latent noise and $\mathbf{c}$ for a conditional target prompt.
\begin{figure}[t!]
\centering
    \includegraphics[width=1.0\textwidth]{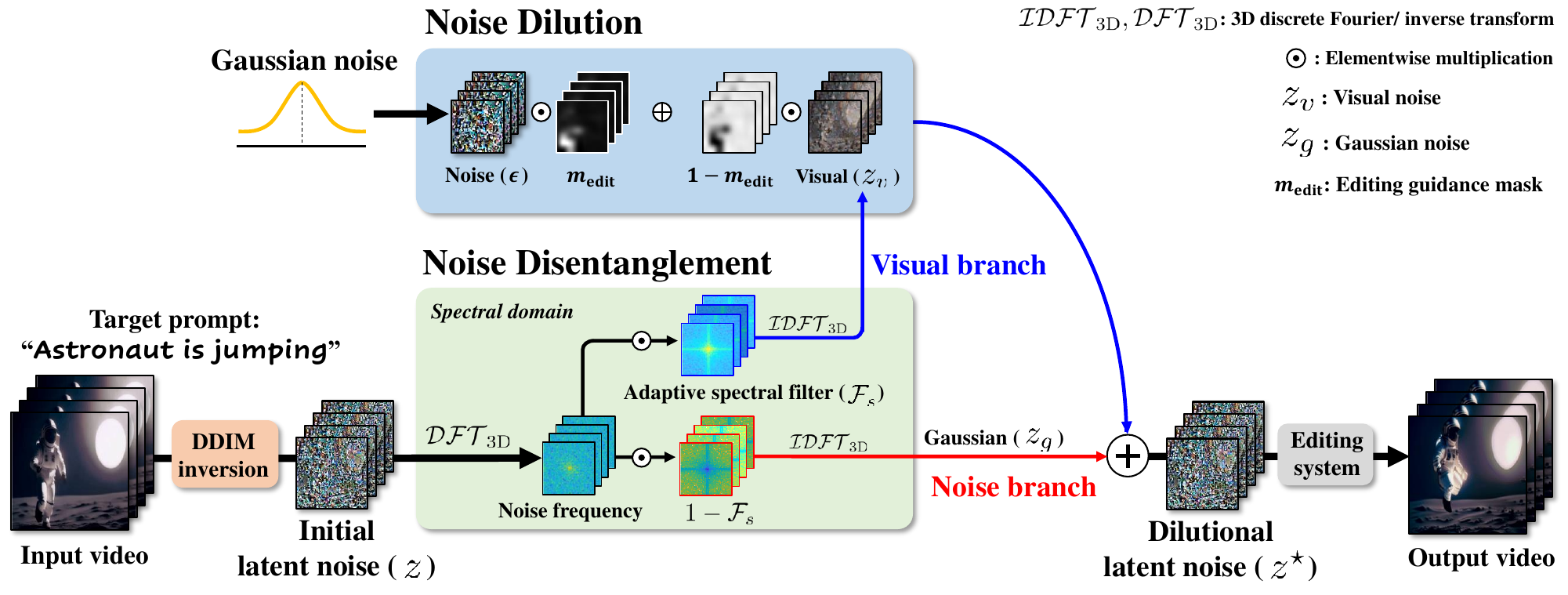}
   \caption{Illustration of Dilutional Noise Initialization (DNI) framework, which refines initial latent noise $z$ into dilutional latent noise $z^{\star}$, enabling editing systems to perform effective editing including non-rigid editing. DNI contains two main modules: (1) Noise Disentanglement which separates the noise $z$ into Gaussian noise $z_{g}$ and visual noise $z_{v}$ containing input video components and (2) Noise Dilution which adds a Gaussian noise $\epsilon$ on the $z_{v}$ to mitigate restrictions of the input video structure near the editing region. The noises $z_{v}$ and $z_{g}$ are recombined to build $z^{\star}$ for an input of video editing.}
\label{fig:model}
\end{figure}
\section{Dilutional Noise Initialization}
Dilutional Noise Initialization (DNI) framework aims to enable video editing systems to perform effective editing including non-rigid modifications.
\cref{fig:model} illustrates the overall process of DNI framework, where it takes initial latent noise $z$ based on $T$ step inverse denoising (\ie DDIM inversion) of the input video and synthesizes dilutional latent noise $z^{*}$ to mitigate constraints from the visual structure of the input video in the area to be edited.
The DNI framework consists of two primary components: (1) Noise Disentanglement and (2) Noise Dilution.
The noise disentanglement separates initial latent noise $z$ into visual noise $z_{v}$ and Gaussian noise $z_{g}$ using our designed adaptive spectral filter in the 3-dimensional frequency domain.
Noise dilution specifies editing region within the visual noise using a target prompt and blends additional Gaussian noise, thereby diminishing the input video's structural influence in the region to be edited.
Finally the dilutional latent noise $z^{\star}$ is synthesized by merging the two noises from the visual branch and the noise branch.
\subsection{Noise Disentanglement}
\begin{wrapfigure}{r}
{0.45\linewidth}
    \centering
    \includegraphics[width=0.45\textwidth]{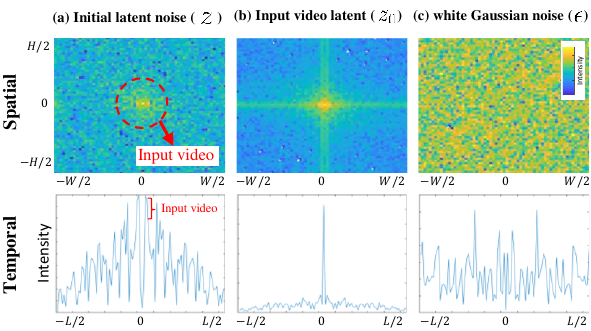}
    \caption{Discrete Fourier transform (DFT) of (a) initial latent noise $z$, (b) video latent feature $z_{0}$, and (c) white Gaussian noise $\epsilon$. A similar distribution between $z$ and $z_{0}$ (red circle) shows that $z$ contains the input video components (top: spatial domain 2D-DFT, bottom: temporal domain 1D-DFT).}
    \label{fig:spectral}
\end{wrapfigure}
Noise disentanglement aims to extract the inherent input video components from the initial latent noise $z$.
To conduct this, we investigated the spectral characteristics of the latent noise across spatial and temporal domains.
\cref{fig:spectral} presents a spatial (top) and temporal (bottom) frequency of (a) initial latent noise $z$, (b) input video latent feature $z_{0}$ prior to noise addition, and (c) white Gaussian noise $\epsilon \sim N(0,I)$ (\ie Isotropic Gaussian).
The low frequency of $z$ shows similar distributions (region in red) with $z_{0}$ in both spatial and temporal frequencies.
Based on this observation, we may apply a low-frequency pass filter (LPF) to acquire components of the input video.
However, the LPF causes the loss of high-frequency components in the input video and also becomes heuristic to correspond to different frequencies for each input video.
Therefore, we introduce an adaptive spectral filter (ASF) that can adaptively respond to the frequency changes in the input video. (\cref{ablation} provides detailed analysis of adaptive spectral filter with visualization in \cref{fig:layoutpassfilter}.)
The ASF builds a frequency pass filter based on the frequency spectrum of the input video latent $z_{0}$, such that it appropriately captures the frequency range of each input video.
Formally, the adaptive spectral filter $\mathcal{F}_{s}$ is defined as below:
\begin{equation}
\mathcal{F}_{s} = \textrm{Norm}_{\textrm{min-max}}(\mathcal{DFT}_{\textrm{3D}}(z_{0})) \in \mathbb{R}^{W \times H \times L \times C},
\end{equation}
where $W,H,L,C$ are the width, height, length, and channel of $z_{0}$. The $\mathcal{DFT}_{\textrm{3D}}$($\cdot$) is 3-dimensional discrete Fourier transform, and $\textrm{Norm}_{\textrm{min-max}}$ is the min-max normalization for the scaling between 0 to 1. 
Thus, employing the $\mathcal{F}_{s}$, we separate initial latent noise $z$ into visual noise $z_{v}$ and Gaussian noise $z_{g}$ as given below:
\begin{equation}
z_{v} = \mathcal{IDFT}_{\textrm{3D}}(\mathcal{F}_{s} \odot \mathcal{DFT}_{\textrm{3D}}(z)), \textrm{  } z_{g} = \mathcal{IDFT}_{\textrm{3D}}((1 - \mathcal{F}_{s}) \odot \mathcal{DFT}_{\textrm{3D}}(z)),
\end{equation}
where the $\mathcal{IDFT}_{\textrm{3D}}$ is inverse $\mathcal{DFT}_{\textrm{3D}}$ and $\odot$ is the elementwise multiplication.
\subsection{Noise Dilution}
Noise dilution aims to enhance the flexibility of modifications by mitigating the constraints imposed by the input video's visual structure in editing region. 
Thus, by taking visual noise $z_{v}$ as input, the noise dilution specifies the editing region on the $z_{v}$ using the target prompt and blends additional Gaussian noise there.
We describe this process as editing guidance and noise blending in the following.
\paragraph{\bf Editing guidance.}
\begin{wrapfigure}{r}
{0.45\linewidth}
    \centering
    \includegraphics[width=0.45\textwidth]{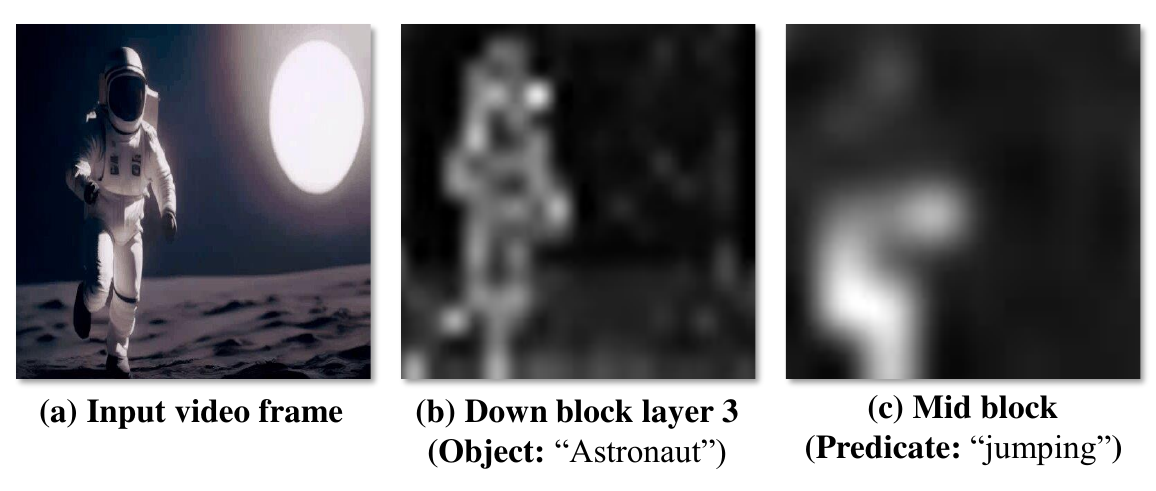}
    \caption{Cross-attention map between frame and reference words.}
    \label{fig:attention}
\end{wrapfigure}
To provide guidance about the editing region, we first select the editing reference words (\eg `Astronaut', `jumping') in the textual prompts (\eg Astronaut is jumping) and obtain the guidance mask $m_{\textrm{edit}}$ about these words.
Our initial choice for the mask generation was to use pre-trained segmentation model (\eg SAM \cite{kirillov2023segment}), but this was not appropriate for specifying the region for non-rigid editing due to noun-based detection. (\ie Non-rigid editing mainly requires motion/pose modification by the predicate in a target prompt.)
Therefore we use a cross-attention map between the reference words and video frames in the diffusion model.
To be specific, as shown in \cref{fig:attention} (b), when the reference word is a noun or adjective, the attention map $m_{\textrm{rgd}}$ from the down-block of the UNet is used to provide the clear boundary of the editing target for rigid modification.
In \cref{fig:attention} (c), when the reference word is a predicate, blurry attention map $m_{\textrm{non-rgd}}$ of mid-block is employed to encompass areas of non-rigid modification.
All attention maps of $m_{\textrm{rgd}}, m_{\textrm{non-rgd}}$ are resized to the latent spatial dimension (\ie $W \times H$) and added together to form an editing guidance mask $m_{\textrm{edit}}$\footnote{Multiple attention maps of $m_\textrm{rgd}$ and $m_\textrm{non-rgd}$ by multiple reference words are mean-pooled before the \cref{eq:3} and $m_{\textrm{edit}}$ is scaled between 0 to 1 after the \cref{eq:3}} as given below:
\begin{equation}
m_{\textrm{edit}} = \alpha \times m_{\textrm{rgd}} + \beta \times m_{\textrm{non-rgd}} \in \mathbb{R}^{W \times H \times L},
\label{eq:3}
\end{equation}
where $\alpha$ and $\beta$ are hyperparameters between 0 and 1 that modulate the mask's intensity. 
If closer to 1, they increase Gaussian noise and suppress the input video's influence through noise blending. 
The $\alpha$ enables effective modifications within the visual structure, while $\beta$ supports editing beyond those structures.
\paragraph{\bf Noise blending.}
To the specified editing region by the mask $m_{\textrm{edit}}$, we blend Gaussian noise to diminish the input video structural influence.
To conduct this, the visual noise $z_{v}$ and white Gaussian noise $\epsilon$ are blended based on the $m_{\textrm{edit}}$ with channel-wise broadcasting.
After that, the separated Gaussian noise $z_{g}$ of the initial noise is also combined to build dilutional latent noise as given below:
\begin{equation}
z^{\star} = z_{g} + m_{\textrm{edit}} \times \epsilon + (1 - m_{\textrm{edit}}) \times z_{v} \in \mathbb{R}^{W \times H \times L \times C}.
\label{eq:comb}
\end{equation}
The dilutional noise $z^{\star}$ is then used for the denoising of video editing systems.
\subsection{Plug-and-play DNI framework}
Recent video editing systems predominantly are largely grouped into two distinct operational approaches: (1) tuning-based \cite{wu2022tune,liu2023video} and (2) tuning-free methods \cite{geyer2023tokenflow,qi2023fatezero}.
The DNI framework can be applied to both methods, offering a model-agnostic approach that enhances editing versatility.
In an inference time (\ie denoising) of a model, DNI injects dilutional latent noise $z^{\star}$ in the model as input instead of initial latent noise $z$ as given below:
\begin{equation}
\mathcal{V}_{\textrm{edit}} = \mathrm{\textrm{Denoise}}(z^{\star},\mathcal{T}),
\end{equation}
where the $\mathcal{T}$ is target prompt and $\mathcal{V}_{\textrm{edit}}$ is the edited video.
\section{Experiment}
\subsection{Experimental Settings}
\paragraph{\bf  Implementation Details.} 
The VQ-VAE \cite{van2017neural} is used for patch-wise frame encoding, and CLIP model (ViT-L/14) \cite{radford2021learning} for text embedding.
We follow original settings of the baselines for video diffusion models (\ie Stable Diffusion v1.5 and v2.1).
Experiment is performed on NVIDIA A100 GPU, where the $W,H$= 64 is used, rigid editing uses 0.3<$\alpha$<0.7, $\beta=0$, and non-rigid editing use $\alpha$<0.4, $\beta$>0.6.
For the tuning-free model, DNI is applied to all injections of latents tuned by a source prompt.
Empirically, we found leveraging the noise branch in Eq. (\ref{eq:comb}) also enhances effectiveness for some edits, so we utilize this by flexibly multiplying $0 < \gamma < 2.0$ to noise branch and $2-\gamma$ to visual branch.
\paragraph{\bf Dataset and Baselines.}
We validate videos on DAVIS \cite{pont20172017} and LOVEU-TGVE \cite{wu2023cvpr}, which are video editing challenge dataset\footnote{https://sites.google.com/view/loveucvpr23/track4} comprising 32 to 128 frames of each. 
DNI framework is validated about non-rigid/rigid editing on recent editing systems including Tune-A-Video (TAV) \cite{wu2022tune}, Video-P2P (VP2P) \cite{liu2023video}, FateZero (FZ) \cite{qi2023fatezero} and TokenFlow (TF) \cite{geyer2023tokenflow} on their public codes.
\subsection{Evaluation Metric}
Editing results are validated using four criteria: (1) textual alignment, (2) input fidelity, (3) frame consistency, and (4) human preference.
The textual alignment measures the semantic alignment between a target prompt and an edited video using the CLIP score \cite{radford2021learning} and PickScore \cite{kirstain2023pick}.
The PickScore approximates human preferences by a large-scale trained model.
The fidelity measures the preservation of original content in the unedited region using learned perceptual image patch similarity (LPIPS), and structural similarity index measure (SSIM).
The frame consistency measures image CLIP scores between sequential frames and measures fréchet video distance (FVD) to evaluate the naturalness of videos.
For the human evaluation, we investigate the preferences of edited videos according to the target prompt between the editing models and the models with DNI.
\begin{figure}[t!]
\centering
    \includegraphics[width=1.0\textwidth]{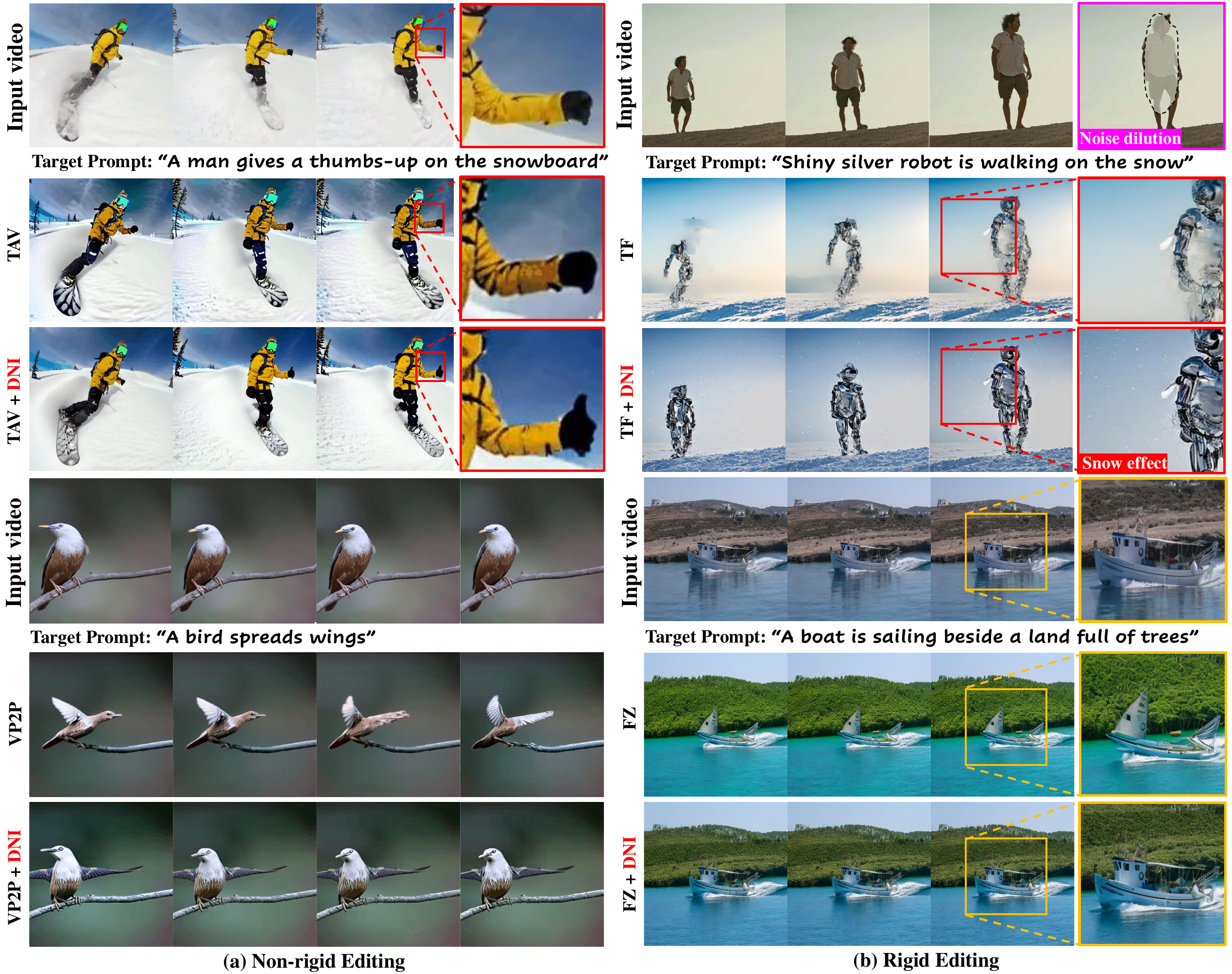}
   \caption{Qualitative results of applying DNI on recent editing systems according to (a) non-rigid editing (motion change) and (b) rigid editing (style transfer, object overlay). TAV: Tune-A-Video, VP2P: Video-P2P, FZ: FateZero, TF: TokenFlow. The yellow box shows a zoomed view of fidelity on unedited regions and the red box shows editing effects. Editing reference words are \{gives, thumbs-up\}, \{spreads, wings\}, \{shiny, silver, robot, snow\}, \{land, trees\} respectively for each sample. The noise dilution in the top of the right (\ie region in black dotted line) is visualized by overlapping together with the input frame to show that the editing based on dilution is seamlessly connected with the surroundings, even to naturally extend the scope of dilution.}
\label{fig:qual}
\end{figure}
\subsection{Experimental Results}
\paragraph{\bf Qualitative Results.}
\cref{fig:qual} presents qualitative results of DNI framework incorporating with recent editing systems \cite{wu2022tune,liu2023video,geyer2023tokenflow,qi2023fatezero}.
To validate qualitative impact of DNI framework, we conduct case studies in terms of two distinct editing categories: (a) non-rigid editing and (b) rigid editing.
For non-rigid editing, current editing systems fail to synchronize with the intended target prompt, resulting in the reconstruction of original input videos or the improper fusion of original contents (\eg trees) with the desired alterations (\eg wings).
However, these models using the DNI effectively perform non-rigid editing on humans and objects.
Remarkably, motion editing for actions such as a thumbs-up (\ie red box) is selectively performed based on the visibility of the skier's hand. 
This precision is attributed to the sensible application of noise dilution, which is selectively applied to the visible editing region (\ie the hand) throughout the video frames.
For rigid editing, including object overlay at the top and style transfer at the bottom, both current editing systems and those enhanced with DNI achieve qualitatively appropriate modifications.
However, upon closer comparison, solely the models utilizing the DNI framework preserve a superior fidelity within the unedited regions (\ie yellow box) of the video.
It is considered that selective dilution by editing the guidance mask improves fidelity in the model.
At the top, we adapted the model to transform a man into a robot walking in the snow.
Intriguingly, the model with the DNI framework not only alters the human's appearance but also adds a snowing effects in the sky.
For this sample, we also visualize the diluted region on top of the input frame (\ie frame outlined by pink color) indicated by a black-dotted line.\footnote{Although dilution is applied over the initial latent noise, we marked an area over the actual frame to help qualitatively understand the dilution effect.}, where it shows that the scope of modification surpasses the initially established diluted perimeters.
This indicates that the editing based on the diluted latent noise is seamlessly connected with the surrounding area and naturally extends the editing effect to the surroundings, enhancing the overall effectiveness of the editing.
\begin{table}[t]
\caption{Evaluations of edited videos from DAVIS and TGVE for non-rigid/rigid type editing in terms of textual alignment, fidelity, consistency, and human preference. $\textrm{CLIP}^{\star}$: text-video clip,
$\textrm{CLIP}^{\dagger}$: image-image clip, P-Score: PickScore, PF: preference}
\centering
\begin{center}
\begin{tabular}{lccccccc} \toprule[1pt]
& \multicolumn{2}{c}{\bf Textual Alignment} &\multicolumn{2}{c}{\bf Fidelity} &\multicolumn{2}{c}{\bf Consistency} & {\bf Human}
\\ 
\cmidrule(lr){2-3} 
\cmidrule(lr){4-5} 
\cmidrule(lr){6-7}
& $\textrm{CLIP}^{\star}$ $\uparrow$ & P-Score $\uparrow$ & $\textrm{LPIPS}$ $\downarrow$ & $\textrm{SSIM}$ $\uparrow$ & $\textrm{CLIP}^{\dagger}$ $\uparrow$ & FVD $\downarrow$  &PF $\uparrow$ \\ \midrule
TAV \cite{wu2022tune}           &22.6/27.1 &19.5/20.2 &0.193/0.181 &0.621/0.653 &0.921/0.952 &3481/3392 &0.14 \\
TAV+DNI                &27.6/28.5 &20.6/20.9 &0.168/0.161 &0.706/0.711 &0.952/0.961 &3270/3151 &0.86 \\ \midrule 
FZ \cite{qi2023fatezero}  &21.2/26.1 &19.4/20.1 &0.173/0.165 &0.636/0.643 &0.958/0.963 & 3319/3106 &0.34 \\ 
FZ+DNI           &26.1/28.7 &20.1/21.2 &0.168/0.157 &0.672/0.687 &0.965/0.968 & 3209/3071 &0.66 \\ \midrule
VP2P \cite{liu2023video}   &22.5/27.2 &19.6/20.0 &0.181/0.172 &0.645/0.677 &0.954/0.958 & 3231/3095 &0.38 \\
VP2P+DNI          &\textbf{27.9}/29.3 &\textbf{20.9}/21.3 &0.161/0.158 &0.717/0.719 &0.961/0.964 &3135/2953 &0.62 \\ \midrule
TF \cite{geyer2023tokenflow}   &21.7/27.4 &19.4/20.1 &0.160/0.157 &0.653/0.677 &0.971/0.974 & 3152/3043 &0.41 \\
TF+DNI          &25.9/\textbf{29.6} &20.7/\textbf{21.5} &\textbf{0.143}/\textbf{0.151} &\textbf{0.731}/\textbf{0.733} &\textbf{0.980}/\textbf{0.977} &\textbf{3103}/\textbf{2912} &0.59 \\ \bottomrule[1pt]
\end{tabular}
\end{center}
\label{mytab:1}
\end{table}
\paragraph{\bf Quantitative Results.}
\cref{mytab:1} provides evaluations of non-rigid and rigid editing on DAVIS and TGVE videos using recent editing systems with the DNI framework. 
The assessments covers textual alignment, fidelity, consistency, and human evaluation.
The effectiveness of the DNI framework is validated across all the video editing systems, with a notable enhancement in textual alignment particularly observed within the realm of non-rigid editing.
For fidelity, it measures the preservation of unedited areas in the video after masking the same specified regions for editing.
The fidelity is lower in the tuning-based models (\ie TAV, Video-P2P) compared to tuning-free models (\ie FateZero, TokenFlow).
Tuning-based models train the diffusion model based on the input video, which sometimes leads to overfitting. 
Therefore, in the inference of editing, overly adhering to the trained video results in unnatural edits.
In these models, dilutional noise identifies areas for additional noise to obscure and reduces over-reliance on the initial visual structure. 
This allows for better adaptation to the target prompt, improving naturalness, fidelity, and consistency.
Tuning-free models excel in consistency but are less responsive to target prompts, making them suitable only for rigid editing due to their dependence on input video structure in the initial latent noise. 
The dilutional noise adaptively reduces this reliance and enhances the effectiveness of editing.
\begin{figure}[t]
\centering
    \includegraphics[width=1.0\linewidth]{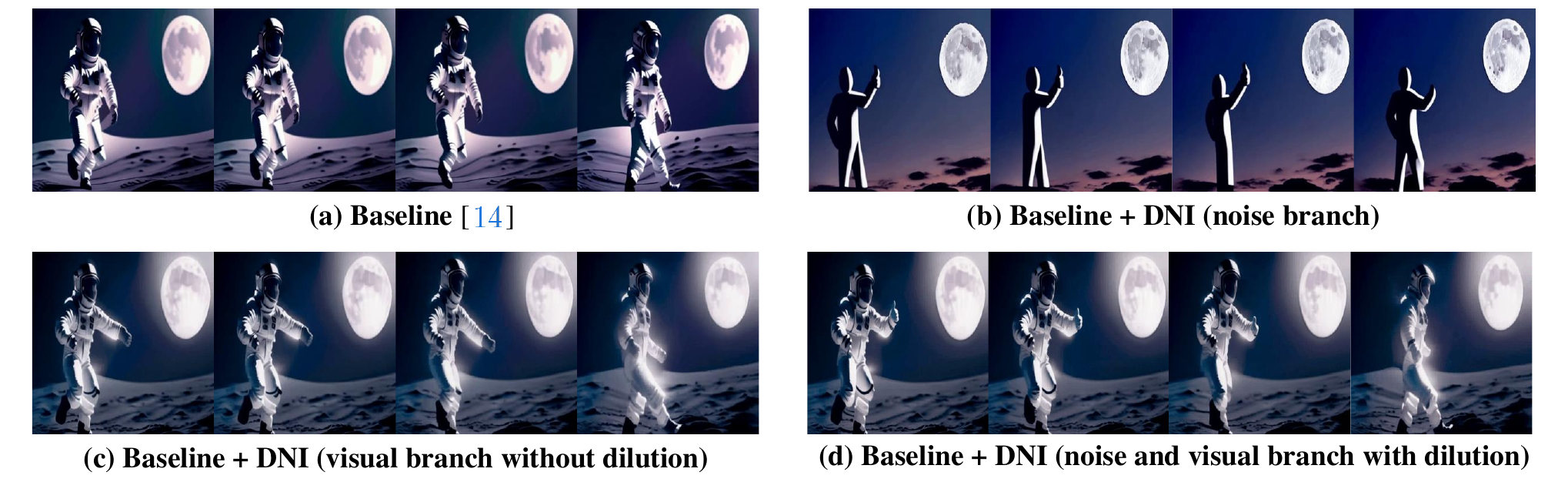}
   \caption{Ablation studies about visual and noise branches in DNI. The input video is shown in \cref{fig:teaser}. The target prompt is ``Astronaut gives a thumbs-up under the moon''.}
\label{fig:ablation_1}
\end{figure}
\subsection{Ablation Study}
\label{ablation}
\paragraph{\bf Ablative results about noise and visual branch.}
To investigate the effectiveness of the visual and noise branches in the editing of DNI framework, in \cref{fig:ablation_1}, we perform ablation studies of these two branches about editing a video based on a target prompt ``Astronaut gives a thumbs-up under the moon''.
The target prompt demands a complex blend of rigid and non-rigid editing, necessitating the adjustment of the hand’s pose and reimaging the pale glow planet behind as moon.
\cref{fig:ablation_1} (a) shows the editing results for the baseline \cite{liu2023video}.
In the edited video, the planet in the sky is changed into the moon, but the hand gesture fails to transition from the original motion to the intended thumbs-up, displaying an unnatural pose between the thumbs-up and its original stance.
\cref{fig:ablation_1} (b) and (c) show the results of integrating the DNI framework using only one of the branches, either the visual branch or the noise branch.
The results of (b) show the edited frames using only the noise branch. 
While effective changes are shown in the results, they significantly deviate from the input video, especially in the original astronaut.
This indicates that the input video's visual structure within the initial latent noise plays a role in maintaining fidelity to the input video.
The outcomes from (c) illustrate the results of employing the visual branch without dilution.
Unlike (b), these results maintain a strong fidelity to the original video, yet they fall short in executing non-rigid editing for a thumbs-up, resulting in edits that display awkward motion through unnatural elongation of the arm.
The results of (d) show the edited video combining the two branches with applying dilution to the visual branch.
It maintains fidelity to the astronaut well, while successfully applying non-rigid editing about thumbs-up.
Notably, when the arm is shaded by the torso, (\ie fourth frame) the thumbs-up also disappears, resulting in a natural movement.
This denotes that dilution is discerningly applied based on the motions displayed within the video frames.
\begin{figure}[t]
\centering
    \includegraphics[width=1.0\linewidth]{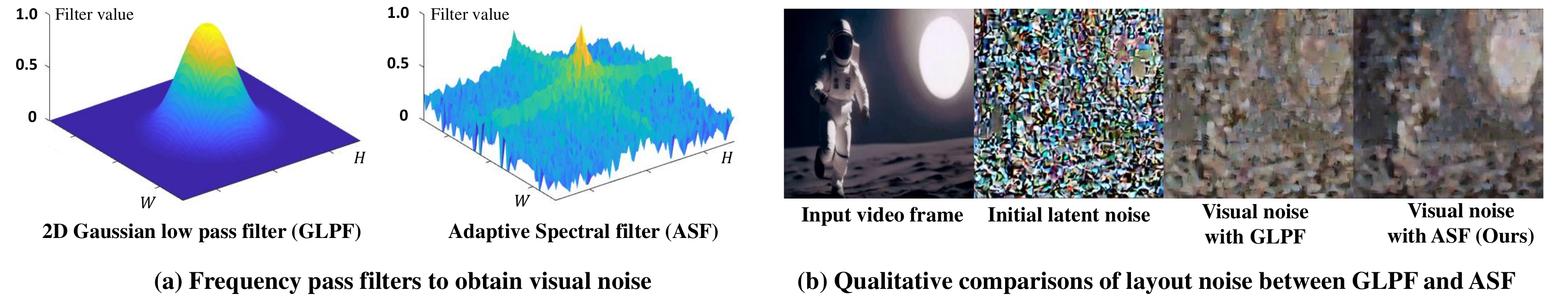}
   \caption{Frequency pass filters for extracting visual noise from initial latent noise. (top-left: Gaussian low pass filter, top-right: our proposed adaptive spectral filter, bottom: frequency-filtered results using GLPF and ASF).}
\label{fig:layoutpassfilter}
\end{figure}
\paragraph{\bf Effectiveness of adaptive spectral filter.}
\begin{wraptable}{r}{0.4\textwidth}
%\scriptsize
\small
  \caption{Quantitative validation to assess the involvement of input video's component within the visual noise, filtered by different frequency pass filters.\\}
  \centering
  \begin{tabular}{l c}
    \toprule
    %\multirow{1}{*}{Model} & B1 & R-1 &R-L \\
    Frequency pass filter type & PSNR \\
\midrule 
    %\Xhline{2\arrayrulewidth}
    Gaussian low pass filter ($\sigma=1)$  & 6.2  \\
    Gaussian low pass filter ($\sigma=3)$ & 9.1 \\
    Gaussian low pass filter ($\sigma=5)$ & 13.2 \\
    Gaussian low pass filter ($\sigma=10)$ & 11.3 \\ \midrule
    adaptive spectral filter (Ours) & \textbf{18.4} \\
    \bottomrule
  \end{tabular}
  \label{tab:LPF_exp}
\end{wraptable}
The initial latent noise fed into video editing systems contains the inherent visual structure of the input video.
Within the DNI framework, noise disentanglement aims to isolate this input video structure from the initial latent noise.
\cref{fig:layoutpassfilter} shows the adaptive spectral filter (ASF) we designed for this purpose.\footnote{Although ASF is a 3D filter, for visual clarity, we show a 2D filter in spatial domain.}
As shown in \cref{fig:spectral}, the input video contains multiple frequencies ranging from low to high, and some of these remain in the initial latent noise.
To appropriately capture these, we use the frequencies of the input video as a frequency filter by scaling between 0 to 1.
To demonstrate the effectiveness of the ASF, we compare it with a Gaussian low pass filter (GLPF) in the spatial domain.
\cref{fig:layoutpassfilter} (b) shows the input video frame, its initial latent noise, and frequency-filtered visual noise using the GLPF and ASF.
Qualitatively, the visual noise obtained using ASF more clearly reveals the input video's contents.
To quantitatively measure this, in  \cref{tab:LPF_exp}, we conducted comparisons in terms of the peak signal-to-noise ratio (PSNR) with the input video and filtered videos using the ASF and variants of GLPF\footnote{It follows the function of $G(x,y) = e^{-(x^2 + y^2)/2\sigma^{2}}$ to scale from 0 to 1.} by adjusting $\sigma$ to leverage filtering frequency band.
This shows that ASF extracts the input video structure more effectively than all variations of the GLPF.
\begin{figure}[t]
\centering
    \includegraphics[width=1.0\linewidth]{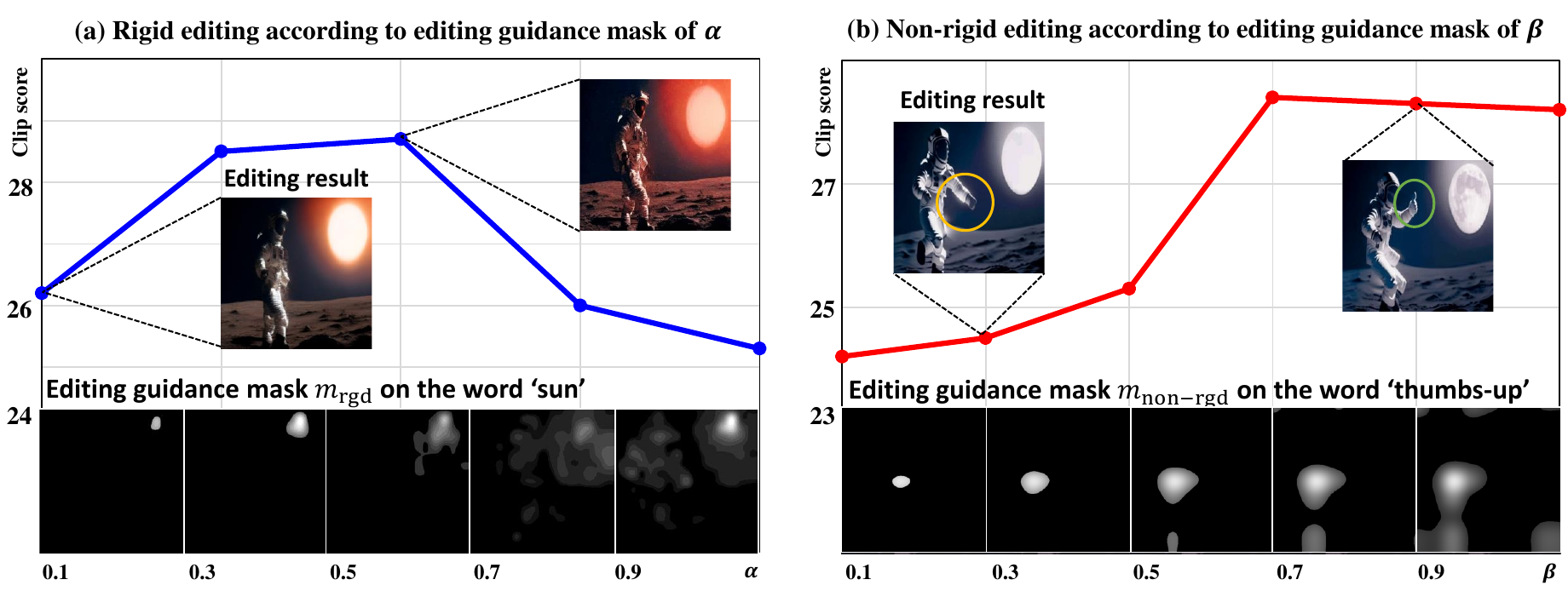}
   \caption{Sensitivity analysis about textual alignment and edited video according to editing guidance mask modulated by $\alpha$ and $\beta$ in \cref{eq:3}. For the setting of rigid editing in (a), the target prompt uses ``Astronaut walks through red sunset'', the parameter $\beta=0$ is fixed, and the reference word for editing is specified as `red' and `sunset', where the mask above visualizes the attention map of `sunset'. For non-rigid editing of (b), the target prompt uses ``Astronaut gives a thumbs-up under the moon'', the $\alpha=0.2$ is fixed, the reference word for editing is specified as `gives', `thumbs-up', and `Astronaut' for (b), where the mask visualizes attention map of `thumbs-up'.}
\label{fig:sensitiviy}
\end{figure}
\paragraph{\bf Sensitivity analysis on editing guidance mask.}
\cref{fig:sensitiviy} presents a sensitivity analysis of the editing results and their textual alignment in response to changes in the editing guidance mask variation.
\cref{fig:sensitiviy} (a) shows the editing results for rigid editing, where `red' and `sunset' are selected as reference words from the target prompt to obtain the editing guidance mask.
Based on our designed \cref{eq:3}, the influence of rigid editing mask $m_{\textrm{rgd}}$ can be adjusted according to variation of mask scaler $\alpha$.
Therefore, we investigate the results of rigid editing according to variation of $\alpha$, and below, the mask $m_{\textrm{rgd}}$ is visualized together. (\ie To enhance the visibility of changes in the attention map, we visualized the map for a single word `sunset'.)
For rigid editing, a 16x16 attention mask is obtained from the down-block of the UNet, which is then resized to 64x64. It can be observed that as $\alpha$ increases, the influence of the mask becomes more intensive amplified by the $\alpha$.
Thus, the edited area also expands, and simultaneously, more effects are incorporated into the video.
This leads to an improvement in textual alignment, and it can also be observed that alignment starts to decline at points where $\alpha$ exceeds 0.7, indicating that excessive effects have been introduced.
\cref{fig:sensitiviy} (b) displays the edited videos and textual alignment for non-rigid editing according to mask scaler $\beta$.
The reference words selected were `gives', `thumbs-up,' and `Astronaut'.
The mask $m_{\textrm{rgd}}$ for `Astronaut' utilizes a 16x16 attention map for the rigid editing,\footnote{Natural language toolkit \cite{loper2002nltk} is used to automatically classify part of speech about words in target prompt, where DNI uses the rigid editing mask for words about objectives and adjectives and the non-rigid editing mask for words in predicate.} and the masks $m_{\textrm{non-rgd}}$ for `gives' and `thumbs-up' employs an 8x8 attention map provided by the mid-block of the UNet to enhance non-rigid editing.
By fixing $\alpha=0.2$, we investigate the variation of non-rigid editing according to the mask scaler $\beta$.
As the $\beta$ increases, it is observed that the highlighted area for non-rigid editing expands, and simultaneously, the thumbs-up, which was not edited (\ie yellow circle) with lower $\beta$, is gradually being synthesized. (\ie green circle).
This indicates that the expansion of dilution by the mask area mitigates the constraints imposed by the input video, allowing the synthesis effect to be free from the original motion of an object (\ie walking motion of arms and hands).
Therefore, $\alpha$ and $\beta$ influence editing effectiveness and also hold editing robustness regions (\ie 0.3<$\alpha$<0.7 for rigid editing and $\beta$>0.6 for non-rigid editing) to properly synthesize desired attributes to conform to the target prompt.
\paragraph{\bf Image editing with DNI framework.}
\begin{wrapfigure}{r}
{0.45\linewidth}
    \centering
    \includegraphics[width=0.45\textwidth]{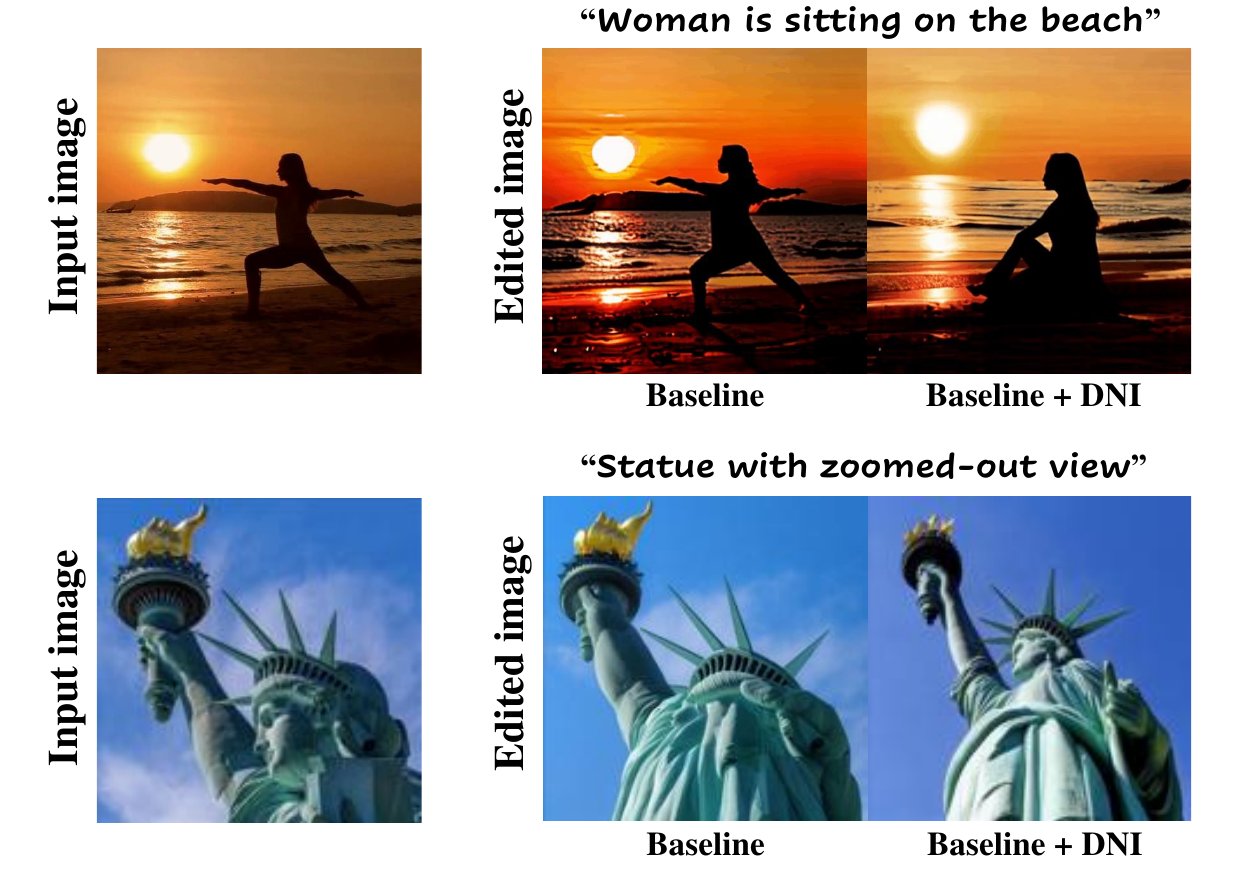}
    \caption{Application DNI framework into image editing system \cite{hertz2022prompt} in terms of non-rigid editing.}
    \label{fig:image_editing}
\end{wrapfigure}
The DNI framework is structurally designed for flexible adaptability within diffusion-based editing systems, such that we apply our work to diffusion-based image editing.
\cref{fig:image_editing} illustrates the enhanced capabilities of the editing system \cite{hertz2022prompt} when incorporated with the DNI framework, showcasing the successful application of non-rigid editing that was previously unattainable by the baseline model.
To be specific, the top of \cref{fig:image_editing} shows non-rigid editing about the pose change, where the current image editing model outputs an image similar to the input image, unable to change the pose according to the target prompt. 
However, when integrated with the DNI framework, it demonstrates the ability to perform precise editing.
The bottom of \cref{fig:image_editing} represents non-rigid editing that changes the view. 
The model, in attempting to switch to a zoomed-out view, is constrained by the layout of the input image and fails to transform correctly.
The results of the incorporation with the DNI framework show that editing can be performed more freely, not restricted by the layout.
\section{Conclusion}
This paper introduces a diffusion-based video editing framework, termed Dilutional Noise Initialization (DNI), designed to facilitate intricate non-rigid modifications of subjects or objects within video.
We introduce a novel concept of `noise dilution' which adds Gaussian noise into initial latent noise to alleviate the restrictive influences imposed by the input video's visual structure on the specified editing regions.
DNI can be easily applied to any diffusion-based editing system in a model-agnostic manner and enhances them to perform non-rigid editing.
Extensive experiments validate its editability and visual effectiveness.
%
%
%
%
%
% TODO REVIEW/FINAL: This \clearpage needs to be removed from both review and camera-ready versions.
\section*{Acknowledgements}
This work was partly supported by Institute for Information \& communications Technology Planning \& Evaluation (IITP) grant funded by the Korea government(MSIT) (No. 2021-0-01381, Development of Causal AI through Video Understanding and Reinforcement Learning, and Its Applications to Real Environments) and partly supported by the National Research Foundation of Korea (NRF) grant funded by the Korea government(MSIT) (No. 2022R1A2C2012706).
\bibliographystyle{splncs04}
\bibliography{main}
\end{document}